\pdfoutput=1
\documentclass[11pt]{article}

\usepackage[preprint]{acl}
\usepackage{times}
\usepackage{latexsym}

\usepackage[T1]{fontenc}
\usepackage[utf8]{inputenc}

\usepackage{microtype}

\usepackage{inconsolata}
\usepackage{multicol}
\usepackage{multirow}
\usepackage{makecell}
\usepackage{graphicx}
\usepackage{amsmath}
\usepackage{array}
\usepackage{tabularray}
\usepackage{xcolor}
\usepackage{soul}

\usepackage{subcaption}
\usepackage{stfloats}
\usepackage{float}
\usepackage{listings}
\usepackage{tabularx}
\usepackage{tikz}

\setul{0.2ex}{0.3ex}
\newcommand{\ulcolor}[2][Red]{\setulcolor{#1}\ul{#2}}

\newcolumntype{L}[1]{>{\raggedright\arraybackslash}p{#1}}
\newcolumntype{M}[1]{>{\centering\arraybackslash}m{#1}}

\definecolor{airforceblue}{rgb}{0.36, 0.54, 0.66}
\definecolor{burgundy}{rgb}{0.5, 0.0, 0.13}
\definecolor{darkolivegreen}{rgb}{0.33, 0.42, 0.18}
\definecolor{goldenrod}{rgb}{0.85, 0.65, 0.13}
\definecolor{brightlavender}{rgb}{0.75, 0.58, 0.89}

\title{CLUE: Concept-Level Uncertainty Estimation for Large Language Models}

\usepackage{authblk}
\author[1]{Yu-Hsiang Wang}
\author[2]{Andrew Bai}
\author[3]{Che-Ping Tsai}
\author[2]{Cho-Jui Hsieh}
\affil[1]{Department of Electrical Engineering, National Taiwan University, Taiwan}
\affil[2]{Department of Computer Science, University of California, Los Angeles}
\affil[3]{Department of Machine Learning, Carnegie Mellon University}
\affil[ ]{\texttt{b08901041@ntu.edu.tw, chepingt@andrew.cmu.edu, \{andrewbai, chohsieh\}@cs.ucla.edu}}

\begin{document}
\maketitle
\begin{abstract}
Large Language Models (LLMs) have demonstrated remarkable proficiency in various natural language generation (NLG) tasks. 
Previous studies suggest that LLMs' generation process involves uncertainty.
However, existing approaches to uncertainty estimation mainly focus on sequence-level uncertainty, overlooking individual pieces of information within sequences. 
These methods fall short in separately assessing the uncertainty of each component in a sequence. 
In response, we propose a novel framework for \textbf{C}oncept-\textbf{L}evel \textbf{U}ncertainty \textbf{E}stimation (CLUE) for LLMs.
We leverage LLMs to convert output sequences into concept-level representations, breaking down sequences into individual concepts and measuring the uncertainty of each concept separately.
We conduct experiments to demonstrate that CLUE can provide more interpretable uncertainty estimation results compared with sentence-level uncertainty, and could be a useful tool for various tasks such as hallucination detection and story generation. 
%CLUE can be used for hallucination detection and has shown superior performance over the baseline approach, achieving a 21\% improvement in macro AUROC.
%Additionally, it exhibits a 33\% higher accuracy in aligning with human judgments compared to sequence-level uncertainty.
%We also demonstrate the utility of our method to explain story generation tasks.
\end{abstract}

\section{Introduction}
\begin{table*}[ht]
    \centering
    \resizebox{\textwidth}{!}{
    \begin{tabular}{>{\centering\arraybackslash}m{0.1\textwidth}|m{0.64\textwidth}|>{\centering\arraybackslash}m{0.2\textwidth} >{\centering\arraybackslash}m{0.21\textwidth}}
    && \multicolumn{2}{c}{Sequence-level uncertainty} \\
    &&& \\[-2.3ex]
    Prompt & Answer the question in one single sentence with details: Who is the founder of Apple? & Sample VRO (\citealp{huang2023look}) & SelfCheckGPT-NLI (\citealp{manakul2023selfcheckgpt}) \\
    \hline
    &&& \\[-2.3ex]
    Output 1 & The co-founder of Apple is Steve Jobs, who, along with Steve Wozniak and Ronald Wayne, \ulcolor[brightlavender]{established the company on April 1, 1976}, \ulcolor[goldenrod]{in Cupertino, California}. & 0.811 & 0.13 \\
    &&& \\[-1.5ex]
    Output 2 & Steve Jobs, along with Steve Wozniak and Ronald Wayne, \ulcolor[brightlavender]{co-founded Apple Inc. in 1976}, revolutionizing the technology industry with \ulcolor[airforceblue]{iconic products like the iPhone and MacBook}. & 0.81 & 0.614 \\ 
    &&& \\[-1.5ex]
    Output 3 & Apple was founded by Steve Jobs, Steve Wozniak, and Ronald Wayne, originating \ulcolor[darkolivegreen]{in a garage in Los Altos}. & 0.822 & 0.904 \\
    &&& \\[-1.5ex]
    Output 4 & \ulcolor[brightlavender]{Apple's inception in 1976} was marked by the collaboration of Steve Jobs, Steve Wozniak, and Ronald Wayne, but \ulcolor[burgundy]{Wayne sold his stake shortly after, missing out on Apple's immense success}. & 0.835 & 0.862 \\
    \hline\hline
    \multicolumn{2}{c|}{} & \multicolumn{2}{c}{Concept-level uncertainty} \\
    \multicolumn{2}{c|}{Extracted concepts} & \multicolumn{2}{c}{CLUE} \\
    \hline
    Concept 1 &	Co-founders of Apple (Steve Jobs, Steve Wozniak, Ronald Wayne) & \multicolumn{2}{c}{0.004} \\
    Concept 2 &	\ulcolor[brightlavender]{Apple's establishment in 1976} & \multicolumn{2}{c}{\textbf{0.175}} \\
    Concept 3 &	\ulcolor[goldenrod]{Location of Apple's establishment (Cupertino, California)} & \multicolumn{2}{c}{\textbf{3.554}} \\
    Concept 4 &	\ulcolor[airforceblue]{Iconic Apple products: iPhone and MacBook} & \multicolumn{2}{c}{\textbf{1.629}} \\
    Concept 5 &	\ulcolor[darkolivegreen]{Origination in a garage in Los Altos} & \multicolumn{2}{c}{\textbf{4.411}} \\
    Concept 6 & \ulcolor[burgundy]{Ronald Wayne's stake sale} & \multicolumn{2}{c}{\textbf{7.965}} \\
    Concept 7 & \ulcolor[burgundy]{Missed opportunity for Ronald Wayne} & \multicolumn{2}{c}{\textbf{6.572}} \\
    \end{tabular}
    }
    \caption{An example of sequence-level and concept-level uncertainty in output sequences generated by LLM. 
    The output sequences may contain both consistent information (co-founders of Apple) and varied details across individual samples. 
    Sequence-level uncertainty (Sample VRO and SelfCheckGPT-NLI) falls short in considering each piece of information separately, therefore the produced uncertainty scores for the whole generated sentence becomes less meaningful. In contrast, 
    by breaking down sequences into concepts, our method effectively captures concepts with high uncertainty (highlighted in colored underline), while still identifying the consistent concept (Co-founers of Apple). 
    %Sequence-level uncertainty indicates the probability of the sequences being different from other sequences, where Sample VRO (\citealp{huang2023look}) considers lexicon similarity and SelfCheckGPT-NLI (\citealp{manakul2023selfcheckgpt}) considers mutual entailment. 
    %Our concept-level uncertainty is determined by the average negative logarithm of the likelihood of a concept being relevant to each output sequence.}
    }
    \label{tab:correct_and_hallucination}
\end{table*}

Large Language Models (LLMs) have demonstrated powerful abilities in generating human-like text and attaining exceptional performance in various Natural Language Processing (NLP) tasks.
Previous studies indicate that the generation process of LLMs involves uncertainty~(\citealp{manakul2023selfcheckgpt}, \citealp{huang2023look}). 
This uncertainty arises from the stochastic nature of the sampling process in LLMs, leading to the generation of different outputs for the same given input. 

Measuring the uncertainty in LLM generation is important, as it can serve as a crucial indicator, offering insights into the reliability or diversity aspects of specific tasks. 
For example, in a question-answering (QA) task, high uncertainty in the model's output could be interpreted as a form of hallucination, deviating from the expectation of producing consistent answers. 
In contrast, in the context of a story generation task, high uncertainty could become a favorable characteristic, contributing positively to the diversity of the generated stories. 
Therefore, understanding and quantifying uncertainty in LLM outputs become essential, allowing for task-specific evaluations and ensuring the desired outcomes in various applications.

Various methods exist for measuring the uncertainty of LLMs' output.
Previous approaches have primarily focused on measuring uncertainty at the sequence level (\citealp{manakul2023selfcheckgpt}, \citealp{huang2023look}), treating an entire generated sequence as a single unit. 
These methods are often used to detect hallucinations by identifying output sequences with high uncertainty.
However, a single sequence may contain multiple pieces of information, each with different uncertainty levels. 
Therefore, these methods encounter the ``information entanglement issue'', where they can only measure the overall uncertainty of an entire sequence. 
This limitation hinders a nuanced evaluation of individual components. 
For example, as illustrated in Table~\ref{tab:correct_and_hallucination}, the output sequence in each sample may include both consistent information and distinct details.
Sequence-level methods fail to discern the uncertainty of each component. 

To address the information entanglement issue, we proposed a framework for \textbf{C}oncept-\textbf{L}evel \textbf{U}ncertainty \textbf{E}stimation (CLUE) for LLMs. 
Concepts represent the fundamental meaning of the text, independent of sequence structure or individual lexicons.
We use LLMs with handcrafted one-shot example to extract comprehensive concepts from the generated output sequences.
Each extracted concept is treated as an independent unit, and its uncertainty is measured separately. 
The extracted concepts are then evaluated by an NLI-based zero-shot text classifier, which assigns the predicted entailment score as the concept score. 
Lastly, the uncertainty is determined by the average negative logarithm of the concept score with respect to each output sequence. 
The details of the framework are presented in Section~\ref{methodology}. 

We demonstrate the effectiveness of CLUE in concept-level hallucination detection and its application as a conceptual diversity metric for story generation. 
Our experimental results validate the assumption that highly uncertain concepts are more likely to be hallucinations in tasks requiring consistent output. 
Furthermore, CLUE demonstrates a 21\% improvement in macro AUROC over the baseline method in detecting hallucinations on QA datasets. 
To evaluate CLUE's efficacy in addressing the information entanglement issue, we compare its accuracy in predicting human judgments with sequence-level methods using Amazon Mechanical Turk (AMT).
The results reveal that it exhibits a 33\% higher accuracy, indicating that our concept-level method better aligns with human judgments and is thus easier for humans to understand. 
We also introduce the utility of CLUE as a conceptual diversity metric for story generation. 

\section{Motivation}
% {\color{teal} (Andrew: I was recently informed that calling this section ``Prelim'' would encourage readers to skip it. Consider changing it to ``motivation'' or ``key observations''.)}
\subsection{Information Entanglement Issue}
Previous sequence-level uncertainty methods are limited to assessing uncertainty for the entire sequence.
Given that paragraph-length sequences encompass vast amounts of information, prior methods primarily focus on sequences of sentence length. 
Nonetheless, even a single sentence can be lengthy and filled with extensive information. 
As shown in Table~\ref{tab:correct_and_hallucination}, a sentence-long sequence may still encompass multiple pieces of information simultaneously. 
Addressing this challenge necessitates breaking sequences down into distinct pieces of information and evaluating their uncertainty individually. 

\subsection{Breaking Down Sequences}
To extract information contained in each sequence, it is essential to break down sequences into their constituent components. 
Various methods exist for sequence breakdown, such as tokenization, named-entity recognition (NER), and syntax tree parsing. 
Different methods lead to varying levels of information. 
For example, tokenization breaks down sequences into tokens, representing the lowest level of information in natural language. 
To enhance generalization ability, we employ LLM prompting to break down sequences into information pieces.
By designing few-shot examples for LLMs, we can easily adjust the information level.
In this paper, we focus on extracting high-level concepts, which effectively capture key meanings or ideas from the given text while disregarding lexical information and sequence structure.
%Further details are provided in Section~\ref{sec:concept_extraction}.

\section{Related Work}
\subsection{Uncertainty Estimation for LLMs}
There are numerous methods to measure uncertainty in LLMs.
From the algorithmic aspect, uncertainty estimation can be categorized into two types: token-based and sampling-based methods. 
Token-based uncertainty relies on the output probabilistic distribution for each token from LLMs (\citealp{NEURIPS2021_e4d2b6e6}, \citealp{kuhn2023semantic}, \citealp{manakul2023selfcheckgpt}, \citealp{zhang2023enhancing}, \citealp{huang2023look}).
These methods directly measure the uncertainty of the generated sequence based on this distribution.
However, they cannot be used for black-box LLMs when the output probabilistic distribution is not available. Further, the output probability is often over-confident and may not reflect the actual uncertainty.  
In contrast, sampling-based uncertainty methods generate multiple samples from the same input prompt and calculate the uncertainty based on these output sequences (\citealp{manakul2023selfcheckgpt}, \citealp{huang2023look}).
For example, \citealp{huang2023look} propose Sample VRO, which is calculated based on the similarity between multiple output samples. 
Sampling-based methods only require output sequences to calculate uncertainty, thereby making them more applicable across a wider range of LLMs.
% In our framework, we focus on sampling-based uncertainty methods.

From the uncertainty level aspect, previous uncertainty methods can be categorized into three levels: sequence-level, token-level, and word-level.
Sequence-level methods treat the entire output sequence as a single unit and assess its uncertainty (\citealp{manakul2023selfcheckgpt}, \citealp{huang2023look}, \citealp{kuhn2023semantic} \citealp{yang2023improving}, \citealp{duan2023shifting}, \citealp{chen2023quantifying}, \citealp{lin2023generating}, \citealp{zhang2023enhancing}, \citealp{hou2023decomposing}, \citealp{rivera2024combining}). 
Notably, most of the sampling-based sequence-level methods can only handle single-sentence sequences. 
Token-level approaches directly measure the uncertainty of individual output tokens (\citealp{tanneru2023quantifying}, \citealp{duan2023shifting}, \citealp{yang2023uncertaintyaware}).
Most of them leverage output token probabilities and employ functions such as entropy or the negative logarithm of the probability for uncertainty estimation. 
Word-level methods involve extracting keywords from output sequences and subsequently evaluating the uncertainty associated with each identified keyword (\citealp{varshney2023stitch}). 
The distinction between word-level and concept-level approaches lies in their functionality. 
Word-level methods only identify keywords present in the output sequences, whereas concept-level methods directly generate concepts based on the key meaning of the output sequence. 

\subsection{Hallucination in LLM Generation}
Hallucination in LLMs refers to the generation of content that deviates from the input prompt or may lack grounding in reality. 
It is important to note that hallucination may present as a factual output but is not relevant to the input prompt. 
Several comprehensive surveys have been conducted to explore hallucination in LLMs (\citealp{huang2023survey}, \citealp{zhang2023sirens}, \citealp{rawte2023survey}, \citealp{ye2023cognitive}).
In order to improve the reliability of LLMs, extensive studies have been dedicated to the detection of hallucinations (\citealp{10.1145/3583780.3614905}, \citealp{bang2023multitask}, \citealp{mündler2023selfcontradictory}, \citealp{chuang2023dola}, \citealp{sadat2023delucionqa}, \citealp{mishra2024finegrained}, \citealp{wang-etal-2023-hallucination}, \citealp{choi2023kcts}, \citealp{forbes2023metric}, \citealp{zhang2023sac3}, \citealp{chen2024inside}). 
Specifically, some approaches leverage the uncertainty in LLMs to identify unreliable content as hallucinations (\citealp{manakul2023selfcheckgpt}, \citealp{varshney2023stitch}, \citealp{zhang2023enhancing}).
Furthermore, numerous studies focus on mitigating hallucinations through self-refinement by LLMs (\citealp{varshney2023stitch}, \citealp{mündler2023selfcontradictory}, \citealp{dhuliawala2023chainofverification}, \citealp{kang2023ever}, \citealp{liang2024learning}, \citealp{ji-etal-2023-towards}, \citealp{guan2023mitigating}, \citealp{feldman2023trapping}). 
In this paper, we focus on utilizing concept-level uncertainty to detect hallucinations that deviate from the input prompt. 

\section{Methodology} \label{methodology}
\begin{figure*}
    \centering
    \includegraphics[width=\textwidth]{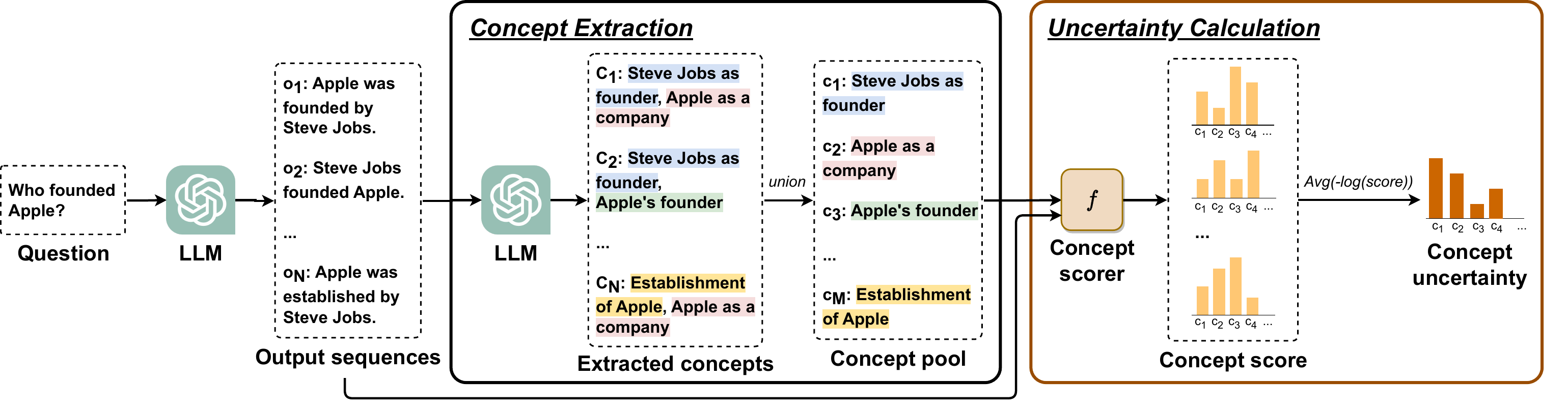}
    \caption{Our proposed framework of concept-level uncertainty. 
    $o_i$ denotes the $i$-th output sequence, $C_i$ denotes the extracted concepts from $o_i$, and $c_i$ denotes the $i$-th concept in the concept pool.}
    \label{fig:framework}
\end{figure*}
We propose a novel framework, CLUE, to measure the uncertainty of LLMs at the concept level. 
CLUE extracts concepts from output sequences in each sample and then assesses concept uncertainty based on the corresponding concept score to each output sequence.
An overview of our framework is presented in Figure~\ref{fig:framework}. 

\subsection{Concept Extraction} \label{sec:concept_extraction}
Concepts are high-level representations of texts, reflecting the meaning of sequences. 
To measure the uncertainty at the concept level, we extract concepts from the generated sequences by prompting LLMs. 
Inspired by \citealp{brown2020language}, we feed handcrafted one-shot example to guide LLMs in generating concepts consistently, as presented in Table~\ref{tab:one_shot_example} in the Appendix.
Our analysis reveals that the length, subject, and quantity of examples barely affect the consistency of extracted concepts. 
% Intriguingly, the use of a one-shot example significantly improves consistency compared to zero-shot prompting. 
We present some examples of generated sequences alongside their corresponding extracted concepts in Table~\ref{tab:seq_and_concepts} in the Appendix. 

We extract a set of concepts for each output sequence. 
Since each output sequence is different, the extracted concepts also vary. 
To comprehensively capture the information that may be generated by the LLM, we combine the sets of concepts extracted from each output sequence to form a unified concept pool. 
The concept pool is composed of the possible concepts generated by the LLM based on the given prompt. 
Since some extracted concepts may exhibit high similarity, we use an NLI-based zero-shot text classifier to automatically consolidate similar concepts, retaining only one instance. 
For example, consider the two closely related concepts: ``Limited competition among ISPs'' and ``Lack of competition in broadband market'', we randomly select one of these concepts to condense the concept pool. 
The zero-shot text classifier is employed to measure the similarity between concepts by computing their mutual entailment scores. 
The two concepts are regarded as equivalent if both entailment scores are higher than the predefined threshold. 
The threshold is set at 0.99 to ensure stringent selection, allowing only very similar concepts to be considered equivalent.
The details of the classifier are presented in Appendix~\ref{appendix:classifier}. 

\subsection{Concept-level Uncertainty Calculation}
\subsubsection{Concept Scorer} \label{sec:concept_scorer}
To measure the concept score based on the relevance between concepts and each output sequence, we design a concept scorer $f$ using an NLI-based zero-shot text classifier. 
Given a sequence $o_i$ and a concept $c_j$, the NLI-based zero-shot text classifier determines whether $o_i$ entails $c_j$ and outputs a probability of entailment.
High entailment probability indicates that $c_j$ is a concept of $o_i$. 
We adopt the entailment probability as the concept score $s_{ij}$. 
The details of the classifier are presented in Appendix~\ref{appendix:classifier}. 
\begin{equation}
    s_{ij} = f(o_i,c_j).
\end{equation}

\subsubsection{Uncertainty Calculation}
We measure the concept score for each concept with respect to each sampled output sequence using the concept scorer. 
The concept uncertainty is determined by calculating the average of the negative logarithm of the concept score
\begin{equation}
  U(c_j) = \text{Avg}_i(-\text{log}(s_{ij})) = -\frac{1}{N}\sum_{i} \text{log}(s_{ij}), 
\end{equation}
where $U(c_j)$ denotes the uncertainty of the concept $c_j$, and $N$ is the number of samples.
Since we employ a sampling-based method for uncertainty calculation, our approach is applicable to both white-box and black-box LLMs.

\section{Experiments}
We conduct experiments on various NLP tasks to demonstrate the utility of the proposed framework. 
In Section~\ref{sec:hallucination_detection}, we illustrate how CLUE detects hallucination at the concept level, which is more intuitive for humans to comprehend compared to sequence-level methods. 
In Section~\ref{sec:diversity_metric}, we extend our framework to another application as a conceptual diversity metric for story generation.

\subsection{Experimental Settings} \label{sec:experimental_setting}
We evaluate the effectiveness of CLUE using question-answering (QA) datasets, which comprise multiple positive and negative instances.
In the context of QA, high uncertainty indicates unpredictability in the generated output. 
Since stability and consistency are expected in QA tasks, high uncertainty implies potential hallucinations.
To prove this statement, we partition the QA datasets into three derivative subsets: the relevant subset $D_R$, the less relevant subset $D_L$, and the irrelevant subset $D_I$. 
$D_R$ and $D_L$ consist of positive and negative instances, respectively, while $D_I$ contains questions paired with answers randomly selected from other instances. 
It is noteworthy that the answers in $D_L$ are more accurate than those in $D_I$, as the incorrect answers of QA datasets are still crafted to respond to the corresponding questions. 
An illustrative example of the distinctions among the three subsets is presented in Table~\ref{tab:dataset_subsets}. 
We subsequently compute the answer concept score $S^a_j$ for each subset to represent the relevance between the answer $a$ and the concept $c_j$ using the Concept Scorer $f$: 
\begin{equation} \label{eq:answer_concept_score}
    S^a_j = f(a, c_j).
\end{equation} 
These answer concept scores then serve as the ground truth for the subsequent evaluation.

\begin{table}[ht]
    \centering
    \resizebox{0.48\textwidth}{!}{
    % \begin{tabular}{m{0.1\textwidth}|m{0.4\textwidth}}
    \begin{tabular}{c|c}
        {} & What county is Farmington Hills, MI in? \\
        \hline
        relevant $D_R$ & \makecell[l]{It is the second largest city in Oakland \\ County in the U.S. state of Michigan.} \\
        \hline
        less relevant $D_L$ & \makecell[l]{In 2010, the area ranked as the 30th \\ safest city in America.} \\
        \hline
        irrelevant $D_I$ & \makecell[l]{The books have since been published \\ by many publishers worldwide.}
    \end{tabular}
    }
    \caption{An example illustrating the three dataset subsets.}
    \label{tab:dataset_subsets}
\end{table}

\subsubsection{Models}
We conduct experiments using OpenAI's GPT-3.5-turbo-instruct model.
During the sampling stage, we set the temperature to 1 and generate $N=5$ samples to produce different outputs while preserving the necessary contextual information for coherent and meaningful responses. 
In the Concept Extraction stage, we set the temperature to 0 to ensure more stable and deterministic results for the extracted concepts.
Additionally, we adopted the NLI-based zero-shot text classifier “bart-large-mnli”~\footnote{\url{https://huggingface.co/facebook/bart-large-mnli}} for our concept scorer. 
It is based on the bart-large model (\citealp{lewis-etal-2020-bart}), pretrained on the MNLI dataset (\citealp{williams-etal-2018-broad}). 

\subsubsection{Datasets}
We select three datasets with different characteristics for a thorough evaluation. 
ELI5-Category is a long-form QA dataset with paragraph-like answers. 
WikiQA consists of simple answers, each sequence comprising only one sentence. 
QNLI is an NLI-based QA dataset that includes answers categorized as either entailing the corresponding questions or not. 
We construct three subsets $D_R$, $D_L$, and $D_I$ for each dataset. 

\paragraph{ELI5-Category} 
The ELI5-Category dataset (\citealp{eli5-category}) is a more recent and compact variant of the original ELI5 dataset (\citealp{fan2019eli5}). 
It is constructed by collecting questions and their answers from \texttt{r/explainlikeimfive} subreddit. 
Each instance contains a single question paired with multiple answers, with each answer being assigned a score.
The score is determined by subtracting the number of downvotes from the number of upvotes given by annotators. 
A higher score indicates a better answer. 
In our experiment, we select answers with the highest and lowest scores for $D_R$ and $D_L$.
As for $D_I$, we randomly choose an answer from another instance to serve as the irrelevant answer. 

\paragraph{WikiQA} 
The WikiQA dataset (\citealp{yang-etal-2015-wikiqa}) consists of 3,047 questions initially sampled from Bing query logs. 
Each instance comprises a single question along with multiple answers, where the answers are sentences extracted from the corresponding Wikipedia page related to the question's topic.
Annotators have labeled each answer as either correct or incorrect. 
In our experiment, we randomly choose one correct answer, one incorrect answer, and one irrelevant answer from another instance to form $D_R$, $D_L$, and $D_I$, respectively.

\paragraph{QNLI} 
The QNLI (Question-answering Natural Language Inference) dataset (\citealp{wang-etal-2018-glue}) is a Natural Language Inference dataset derived from the Stanford Question Answering Dataset v1.1 (SQuAD) (\citealp{rajpurkar-etal-2016-squad}).
Each instance consists of a question associated with a sentence labeled either as ``entailment'' or ``not entailment''. 
In our experiment, we select instances with ``entailment'' sentences as $D_R$ and those with ``not entailment'' sentences as $D_L$. 
For $D_I$, we arbitrarily choose a sentence from another instance as the answer. 
% Notably, the QNLI dataset is a component of the GLUE benchmark (\citealp{wang-etal-2018-glue}).\\

\subsection{Uncertainty-based Concept-level Hallucination Detection} \label{sec:hallucination_detection}
To demonstrate the application of our method for concept-level hallucination detection, we first validate the assumption that high uncertainty in output suggests hallucination. 
Building upon this assumption, we evaluate the effectiveness of CLUE in detecting hallucinations.
We further conduct a human study showing that concept-level uncertainty is better than previous sequence-level uncertainty as it is easier for humans to understand.
\subsubsection{Motivating Experiment}
\begin{table*}
\centering
\begin{tabular}{c|c|c}
\multicolumn{2}{c|}{\textbf{Dataset}} & \textbf{Pearson Correlation} \\
\hline
\multirow{3}{*}{Eli5-Category} & relevant $D_R$ & \textbf{-0.425} \\
& less relevant $D_L$ & -0.374 \\ 
& irrelevant $D_I$ & -0.079 \\
\hline
\multirow{3}{*}{Wiki-QA} & relevant $D_R$ & \textbf{-0.488} \\
& less relevant $D_L$ & -0.33 \\
& irrelevant $D_I$ & -0.062 \\
\hline
\multirow{3}{*}{QNLI} & relevant $D_R$ & \textbf{-0.488} \\
& less relevant $D_L$ & -0.284 \\
& irrelevant $D_I$ & -0.092 \\
\end{tabular}
\caption{Correlation across the three different subsets of the datasets. As expected, the relevant subset has the lowest correlation and the irrelevant subset has the highest correlation. This pattern validates our assumption that concepts with high uncertainty tend to be hallucinated concepts.}
\label{tab:correlation} 
\end{table*}
To verify the assumption that high uncertainty in outputs suggests hallucination, we examine the correlation between the concept uncertainty $U(c_j)$ and the answer concept score $S^a_j$ across all concepts for each instance.
We then compute the average correlation across all instances for three dataset subsets $D_R$, $D_L$, and $D_I$. 
Since the answer concept score indicates the relevance between the concept and the answer, a low correlation implies that concept uncertainty can serve as an indicator of the concept's irrelevance to the answer. 
In $D_R$, where answers are logically connected to the questions, concepts irrelevant to the answer are considered hallucinations. 
We expect a low correlation if the assumption holds. 
Conversely, in $D_I$, where answers are randomly selected from other instances, the answer concept score is not expected to exhibit a clear linear relationship with uncertainty. 
Therefore, we anticipate the correlation for $D_I$ to approach 0.
Regarding $D_L$, the correlation is expected to fall between that of $D_R$ and $D_I$, given its intermediary relevance to the questions.

We present the experiment results of Pearson correlation between concept uncertainty and answer concept score in Table~\ref{tab:correlation}. 
As expected, across three subsets of the datasets, the correlation trend adheres to the following pattern: $D_R$ exhibits a lower correlation than $D_L$, and $D_L$ shows a lower correlation than $D_I$. 
The results demonstrate that across QA datasets with various characteristics, they consistently validate our assumption that concepts with high uncertainty tend to be hallucinated concepts. 
This suggests that the uncertainty of the LLM is an effective measure for assessing the faithfulness of the output across diverse circumstances.
We present an example of the correlation experiment in Table~\ref{tab:corr_examples} in the Appendix.

\subsubsection{Concept-level Hallucination Detection}
\begin{table*}
\centering
\begin{tabular}{c|c|c c c c}
\multirow{2}{*}{\textbf{Dataset}} & \multirow{2}{*}{\textbf{Method}} & \multicolumn{2}{c}{\textbf{Macro}} & \multicolumn{2}{c}{\textbf{Micro}}\\
&& \textbf{AUROC}\ & \textbf{AUPRC} & \textbf{AUROC}\ & \textbf{AUPRC} \\
\hline
\multirow{2}{*}{Eli5-Category} & CLUE & \textbf{0.871} & \textbf{0.894} & \textbf{0.795} & \textbf{0.826} \\
& bart-large-mnli & 0.661 & 0.705 & 0.602 & 0.622 \\
\hline
\multirow{2}{*}{Wiki-QA} & CLUE & \textbf{0.881} & \textbf{0.911} & \textbf{0.838} & \textbf{0.877} \\
& bart-large-mnli & 0.712 & 0.748 & 0.677 & 0.701 \\
\hline
\multirow{2}{*}{QNLI} & CLUE & \textbf{0.867} & \textbf{0.899} & \textbf{0.841} & \textbf{0.884} \\
& bart-large-mnli & 0.761 & 0.798 & 0.76 & 0.789 \\
\end{tabular}
\caption{Experiment results of concept-level hallucination detection. CLUE consistently outperforms bart-large-mnli model across all datasets, showcasing substantial superiority in performance.}
\label{tab:classification}
\end{table*}
Based on the assumption that high uncertainty in outputs suggests hallucination, we proceed to evaluate the efficacy of uncertainty in detecting hallucination. 
We formulate this as a classification task and use concept uncertainty to conduct classification.
%We regard the issue as a classification task, using concept uncertainty as logits to classify concepts. 
To achieve this, we first construct a concept set to be classified, and the label of each concept is determined by its answer concept score, as illustrated in Equation~\ref{eq:answer_concept_score}. 
To enhance precision in concept labeling, we employ two thresholds, a high threshold $\theta_h$ and a low threshold $\theta_l$, applied to the concept scores to determine the concept labels:
\[
    \text{label of concept $c_j$} = 
    \begin{cases}
        0 & \text{if $S^a_j > \theta_h$} \\
        1 & \text{if $S^a_j < \theta_l$} \\
        -1 & \text{otherwise}.
    \end{cases}
\]
A concept is categorized as an ``entailed concept'' (label 0) if its score surpasses the threshold $\theta_h$. 
Conversely, if the score falls below $\theta_l$, the concept is designated as a ``hallucinated concept'' (label 1). 
For this experiment, we do not consider other concepts (label -1). 
We exclusively apply this task on $D_R$ since we require accurate answers from positive instances to label concepts.

As for the metrics, we employ AUPRC (Area Under Precision-Recall Curve) along with AUROC (Area Under the Receiver Operating Characteristic Curve) to evaluate the classification performance. 
Given that each instance contains a concept pool with multiple concepts to be classified, it can be viewed as an independent classification task. 
We present both macro and micro versions of these two metrics to provide insights into the overall performance across all classifications.
Additionally, we compare CLUE to the NLI-based zero-shot classifier “bart-large-mnli” to demonstrate the efficacy of our approach.
The details of the classifier are presented in Appendix~\ref{appendix:classifier}.

The results of the concept-level hallucination detection experiment are presented in Table~\ref{tab:classification}. 
CLUE achieves remarkable performance, significantly outperforming the baseline method in detecting hallucinations. 
Due to the disparity in units between our method and sequence-level uncertainty, direct comparisons of hallucination detection performance with previous methods are not feasible. 
Table~\ref{tab:correct_and_hallucination} provides an example to illustrate that the primary issue with sequence-level uncertainty lies not in its performance but in its unit. 
The ablation studies on the thresholds of concept scores are presented in Appendix~\ref{appendix:ablation}.

\subsubsection{Human Study}
To show that concept-level uncertainty is easier for humans to comprehend, we conduct an experiment directly comparing it with sequence-level uncertainty through human evaluation. 
We generate 100 instances, each comprising a question, along with 2 output sequences and 2 extracted concepts. 
One sequence and concept exhibit high uncertainty, while the other sequence and concept demonstrate low uncertainty.
We treat this task as a binary classification problem and assess the accuracy of using uncertainty to predict the irrelevant option. 
We employ SelfCheckGPT-NLI (\citealp{manakul2023selfcheckgpt}) as the sequence-level method for comparison. 
The instances are labeled using Amazon Mechanical Turk (AMT), where MTurkers are asked to select the concept and sequence they deem more relevant to the given question, as presented in Figure~\ref{fig:interface_seq} and Figure~\ref{fig:interface_concept} in the Appendix. 
To ensure the reliability of human annotations, we assign five distinct MTurkers to each instance. 
The label of each instance is determined based on the option selected by the majority of the MTurkers, i.e. more than 2.  

\begin{table}[h]
\centering
\begin{tabular}{c|c}
\textbf{Uncertainty Method} & \textbf{Accuracy}\\
\hline
% Concept-level & \textbf{0.91} \\
CLUE & \textbf{0.91} \\
\hline
% Sequence-level & 0.58 \\
SelfCheckGPT-NLI & 0.58 \\
\end{tabular}
\caption{Accuracy comparison between concept-level and sequence-level uncertainty in the MTurk experiment. Our approach aligns more closely with MTurkers' judgments.}
\label{tab:mturk}
\end{table}

The results are presented in Table~\ref{tab:mturk}. 
Our concept-level method exhibits a 33\% higher accuracy compared to the sequence-level approach. 
Our findings indicate that concept-level uncertainty correlates more closely with MTurkers' judgments. 
This suggests that CLUE serves as a more effective indicator of the relevance of generated information to the question.

\subsection{Conceptual Diversity Metric for Story Generation} \label{sec:diversity_metric}
As detailed in Appendix~\ref{appendix:previous_diversity_metric}, previous diversity metrics fall short in capturing high-level features such as tone or genre of generated stories.
In this section, we extend the application of our framework to serve as a conceptual diversity metric in story generation. 
\subsubsection{Method}
Since uncertainty cannot directly be used to represent diversity, we define a two-level concept structure: an upper-level concept representing a conceptual feature of generated stories, with lower-level concepts as its subclasses.
For example, consider the overarching concept ``tone'', which includes more specific sub-concepts like ``happy tone'', ``sad tone'', ``humorous tone'', and so forth. 
We measure the diversity of the upper-level concept by aggregating the uncertainty of its lower-level concepts. 
Given that high uncertainty in lower-level concepts indicates that fewer generated stories are considered as the same subclasses, the aggregated uncertainty of lower-level concepts can be regarded as the diversity of the upper-level concept. 
We further propose two aggregation functions: the harmonic mean and entropy. 
The former directly measures the harmonic mean of the uncertainty of all lower-level concepts, while the latter treats it as a multi-class classification problem and measures the entropy of the classes. 
The equations are listed below:
\begin{equation}
    \text{Harmonic mean} = \frac{M}{\sum_{j=1}^M \frac{1}{U(c_j)}}, 
\end{equation}
\begin{equation}
    \text{Entropy} = -\sum_{j=1}^M \frac{n(c_j)}{N}log(\frac{n(c_j)}{N}), 
\end{equation}
where $c_j$ denotes the $j$-th lower-level concept in this experiment, $N$ is the number of samples, $M$ is the number of concepts, and $n(c_j)$ is the number of samples classified as $c_j$:
\begin{equation}
    n(c_j) = \sum_{i} \arg\max_{k} (f(o_i, c_k)) * \delta_{jk},
\end{equation}
\begin{equation}
    \delta_{jk} = 
    \begin{cases}
        1 & \text{if $j = k$} \\
        0 & \text{otherwise}.
    \end{cases}
\end{equation}
\subsubsection{Qualitative Analysis}
To illustrate, we create 1000 stories by prompting LLMs to generate stories with a happy tone. 
We define a set of two-level concepts with an upper-level concept ``tone'' and 5 lower-level concepts ``happy tone'', ``sad tone'', ``humorous tone'', ``serious tone'', and ``romantic tone''. 
As depicted in Table~\ref{tab:diveristy_illustration}, the concept scorer effectively identifies the stories with a happy tone, resulting in significantly lower uncertainty compared to the other lower-level concepts. 
Consequently, in the harmonic mean function, the low uncertainty term predominates in the denominator, leading to low diversity. 
We further create datasets with different diversity to evaluate our metrics. 
The experimental details are listed in Appendix~\ref{appendix:diversity_metric_evaluation}.
\begin{table}[h]
\centering
\begin{tabular}{c|c c}
\textbf{Lower-level Concept} & \multicolumn{2}{c}{\textbf{Uncertainty}} \\
\hline
Happy tone & \multicolumn{2}{c}{\textbf{0.037}} \\
Sad tone & \multicolumn{2}{c}{7.216} \\
Humorous tone & \multicolumn{2}{c}{0.284} \\
Serious tone & \multicolumn{2}{c}{2.949} \\
Romantic tone & \multicolumn{2}{c}{0.241} \\
\end{tabular}
\caption{Uncertainty of the lower-level concepts. Our concept uncertainty score can successfully identify the ground truth (Happy tone). }
\label{tab:diveristy_illustration}
\end{table}

\section{Conclusion}
In this paper, we propose a novel framework for Concept-Level Uncertainty Estimation (CLUE) for LLMs. 
Our framework separates sequences into multiple concepts and measures their uncertainty individually, successfully addressing the information entanglement issue. 
We showcase the versatility of our framework by applying it to hallucination detection and as a conceptual diversity metric for story generation.
We hope the proposed concept-based approach can achieve a more ``interpretable'' uncertainty estimation and can facilitate the interaction between human and LLMs.  
%
%Our experiment results demonstrate the remarkable capability of CLUE in detecting hallucinations, which is proven to be more comprehensible for humans compared to sequence-level uncertainty. 
%We also showcase that CLUE effectively captures the diversity of high-level features of generated stories.

\section*{Limitations}
First, a key limitation of CLUE is its dependency on the chosen LLM for concept extraction and the specific concept scorer utilized.
In this work, we generate a prompt with a one-shot example to improve the consistency of concept extraction. 
In future work, we will explore employing alternative white box methods for concept extraction to enhance the reliability of our framework.

Second, the lack of high-level feature diversity metrics for story generation prevents us from benchmarking CLUE's performance.
However, given the customizable nature of our framework's two-level concept structure, it remains applicable across more scenarios.
In future work, we aim to propose a benchmark for high-level feature diversity measurement in story generation, with CLUE serving as the baseline. 

\section*{Ethical Consideration}
We propose a framework for LLMs to estimate the concept-level uncertainty of generated content. 
The method is designed to improve LLMs' interpretability and improve human-LLM interactions. However, we do believe there could be certain risks if human over-trust the proposed uncertainty estimation tool. For example, there could be implicit biases in LLMs so that the generated biased content will be associated with low uncertainty. Therefore, when using the uncertainty estimation tool, we need to keep in mind that the estimation is measuring the LLM-generated uncertainty, not the true uncertainty of a particular concept. On the other hand, it is also possible that uncertainty estimation is manipulated by adversarial attacks, and further studies are required to improve the robustness of uncertainty estimation against those attacks.  
%Consequently, if LLMs tend to produce toxic content, our method may have the risk of attributing low uncertainty (high reliability) to such content.

% \section*{Acknowledgements}

% This document has been adapted
% by Steven Bethard, Ryan Cotterell and Rui Yan
% from the instructions for earlier ACL and NAACL proceedings, including those for 
% ACL 2019 by Douwe Kiela and Ivan Vuli\'{c},
% NAACL 2019 by Stephanie Lukin and Alla Roskovskaya, 
% ACL 2018 by Shay Cohen, Kevin Gimpel, and Wei Lu, 
% NAACL 2018 by Margaret Mitchell and Stephanie Lukin,
% Bib\TeX{} suggestions for (NA)ACL 2017/2018 from Jason Eisner,
% ACL 2017 by Dan Gildea and Min-Yen Kan, 
% NAACL 2017 by Margaret Mitchell, 
% ACL 2012 by Maggie Li and Michael White, 
% ACL 2010 by Jing-Shin Chang and Philipp Koehn, 
% ACL 2008 by Johanna D. Moore, Simone Teufel, James Allan, and Sadaoki Furui, 
% ACL 2005 by Hwee Tou Ng and Kemal Oflazer, 
% ACL 2002 by Eugene Charniak and Dekang Lin, 
% and earlier ACL and EACL formats written by several people, including
% John Chen, Henry S. Thompson and Donald Walker.
% Additional elements were taken from the formatting instructions of the \emph{International Joint Conference on Artificial Intelligence} and the \emph{Conference on Computer Vision and Pattern Recognition}.

\bibliography{anthology,anthology_p2,custom}

\appendix

\begin{table*}[!t]
    \centering
    \begin{tikzpicture}
    \node (table) [inner sep=4pt] {
    \begin{tabularx}{0.95\textwidth}{X}
        Extract high-level concepts like the following example: \\
        paragraph: ``Basketball, a beloved sport worldwide, has come a long way since its humble beginnings in the late 19th century. The game was originally created by Dr. James Naismith in 1891 as a way to keep his students active during the winter months. Back then, players used a soccer ball and peach baskets as makeshift goals. Fast forward to the modern era, and basketball has transformed into a high-paced, adrenaline-pumping spectacle. With legendary athletes like Michael Jordan, LeBron James, and Kobe Bryant gracing the courts, and the introduction of the slam dunk, three-point shot, and shot clock, the sport has evolved into an art form that captivates fans around the globe. The NBA, with its star-studded roster and global reach, is a testament to basketball's enduring popularity and its remarkable journey from humble beginnings to a multimillion-dollar industry.'' \\
        concepts:``'Basketball's origins', 'Evolution of basketball', 'Modern era of basketball', 'Legendary basketball athletes', 'Basketball's global popularity', 'Basketball as an art form', 'Basketball as a multimillion-dollar industry''' \\\\
        paragraph: \textit{TARGET\_OUTPUT\_SEQUENCE} \\
        concepts: \\
    \end{tabularx}};
    \draw [rounded corners=.5em] (table.north west) rectangle (table.south east);
    \end{tikzpicture}
    \caption{The used prompt with the handcrafted one-shot example.}
    \label{tab:one_shot_example}
\end{table*}

\begin{table*}[!t]
    \centering
    % \begin{tabular}{c c}
    \begin{tabular}{m{0.57\textwidth}|m{0.37\textwidth}}
        \textbf{Output sequences} & \textbf{Extracted concepts} \\
        \hline
        Saliva production increases when we yawn to keep the mouth lubricated and prevent dryness. When we yawn, the muscles in our face and throat contract, causing the movement of saliva and sometimes causing it to shoot out of our mouths. This is a normal and harmless bodily reaction. & ``Saliva production and yawning'', ``Purpose of saliva production during yawning'', ``Muscles involved in yawning'', ``Saliva shooting out during yawning'', ``Normal bodily reaction'' \\\hline
        The sun does not get oxygen from space. The sun is primarily composed of hydrogen and helium, and the process of nuclear fusion creates the energy and light that we see as sunlight. It does not rely on oxygen for this process. & ``Composition of the sun'', ``Nuclear fusion in the sun'', ``Sunlight as a product of nuclear fusion'', ``Sun's energy source'', ``Sun's lack of reliance on oxygen''
        \\\hline
        When we are asleep, our eyes are still functioning, but our brain switches off our ability to send visual signals to the conscious mind. This means that while we may still react to bright lights or movements in our sleep, we are not consciously seeing in the same way as we do while awake. & ``Sleep and vision'', ``Brain activity during sleep'', ``Consciousness and visual signals'', ``Reactions during sleep'', ``Difference between awake and asleep vision''
        \\\hline
        Darkness is the absence of light, and cannot be directly measured as a physical quantity. Light can be measured using units such as lumens or lux, but darkness cannot be quantified in the same way. & ``Darkness' as the absence of light'', ``Measurement of light'', ``Inability to measure darkness''
        \\
    \end{tabular}
    \caption{Examples of generated output sequences and their corresponding extracted concepts.}
    \label{tab:seq_and_concepts}
\end{table*}

\begin{table*}[ht]
    \centering
    % \begin{tabular}{c c c}
    \begin{tabular}{|L{0.1\textwidth} L{0.05\textwidth} m{0.15\textwidth} |M{0.15\textwidth}M{0.1\textwidth} M{0.1\textwidth} M{0.1\textwidth}|}
        \hline
        \textbf{Question} && \multicolumn{5}{L{0.75\textwidth}|}{What does the name ``Meister'' mean in German?} \\\hline
        \rule{0pt}{2.5ex} \textbf{Answer} & $\mathbf{D_R}$ & \multicolumn{5}{|L{0.75\textwidth}|}{Meister means master in German (as in master craftsman, or as an honorific title such as Meister Eckhart).} \\
        \rule{0pt}{2.5ex} & $\mathbf{D_R}$ & \multicolumn{5}{|L{0.75\textwidth}|}{Many modern day German police forces use the title Meister.} \\
        \rule{0pt}{2.5ex} & $\mathbf{D_R}$ & \multicolumn{5}{|L{0.75\textwidth}|}{A rocket engine, or simply ``rocket'', is a jet engine that uses only stored propellant mass for forming its high speed propulsive jet.} \\\hline\hline
        &&& \multicolumn{4}{M{0.35\textwidth}|}{\textbf{Answer concept score}} \\
        \multicolumn{3}{|L{0.4\textwidth}|}{\textbf{Concept}} & \textbf{Uncertainty} & $\mathbf{D_R}$ & $\mathbf{D_L}$ & $\mathbf{D_I}$ \\\hline
        \multicolumn{3}{|L{0.4\textwidth}|}{Skill and expertise associated with the name Meister} & 0.01 & 0.978 & 0.419 & 0.003 \\
        \multicolumn{3}{|L{0.35\textwidth}|}{German origin of the name Meister} & 0.018 & 0.976 & 0.732 & 0.003 \\
        \multicolumn{3}{|L{0.35\textwidth}|}{German cultural influence} & 0.062 & 0.896 & 0.897 & 0.005 \\
        \multicolumn{3}{|L{0.35\textwidth}|}{Achievement and recognition} & 0.302 & 0.525 & 0.889 & 0.037 \\
        \multicolumn{3}{|L{0.35\textwidth}|}{Origin of Meister} & 0.766 & 0.226 & 0.294 & 0.012 \\
        \multicolumn{3}{|L{0.35\textwidth}|}{Definition of Meister} & 0.876 & 0.251 & 0.474 & 0.012 \\
        \multicolumn{3}{|L{0.4\textwidth}|}{Use of Meister as a surname} & 1.371 & 0.073 & 0.471 & 0.005 \\\hline
        \multicolumn{3}{|L{0.4\textwidth}|}{\textbf{Pearson Correlation}} && -0.954 & -0.529 & 0.041 \\\hline
    \end{tabular}
    \caption{An example of the correlation between concept uncertainty and answer concept score across three dataset subsets. In this instance, the concept uncertainty effectively represents the concepts' relevance to the question, resulting in a low correlation for $D_R$ and $D_L$. In the case of $D_I$, where the answer is not accurate to the question, the uncertainty does not exhibit a linear relationship with the answer concept scores.}
    \label{tab:corr_examples}
\end{table*}

\begin{figure*}[ht!]
    \centering
    \includegraphics[width=\textwidth]{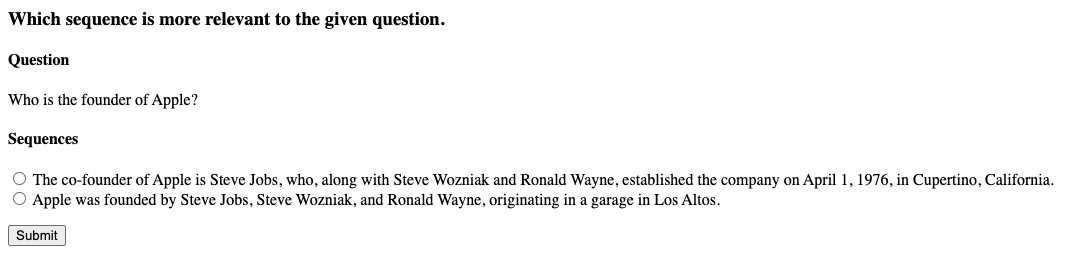}
    \caption{Screenshot of sequence-level human annotation interface presented to MTurkers.}
    \label{fig:interface_seq}
\end{figure*}
\begin{figure*}[ht!]
    \centering
    \includegraphics[width=\textwidth]{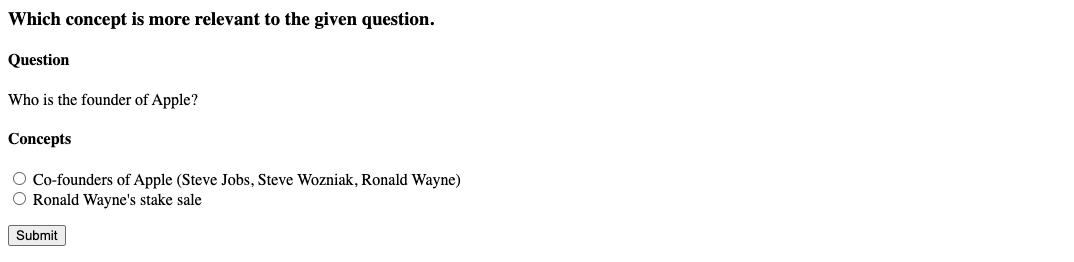}
    \caption{Screenshot of concept-level human annotation interface presented to MTurkers.}
    \label{fig:interface_concept}
\end{figure*}

\section{NLI-based Zero-shot Text Classifier} \label{appendix:classifier}
An NLI-based zero-shot text classifier operates by predicting three logits, each representing the degree of the relationship between the premise and the hypothesis for the labels: ``entailment'', ``contradiction'' and ``neutral''. 
Following the instructions from the bart-large-mnli website, we disregard the ``neutral'' label and apply a softmax layer to the remaining two logits to derive the probability associated with the ``entailment'' label:
\begin{equation}
\begin{split}
    f(o_i,c_j) = \sigma(\text{cls}(o_i,c_j)==\textit{entailment})
\end{split}
\end{equation}
\begin{equation}
    \sigma(x_{i}) = \frac{\exp(x_i)}{\sum_j \exp(x_j)}
\end{equation}
where $\sigma$ denotes softmax function and cls denotes the classifier. 

In our framework, we employ the NLI-based zero-shot text classifier for three purposes: concept consolidation, the concept scorer, and the baseline method for the hallucination detection task.
For concept consolidation, the classifier computes the mutual entailment score of extracted concepts as their similarity. 
One concept serves as the premise, and the hypothesis is generated by transforming the other concept into the following format: ``This concept is similar to {\textit{PREMISE\_CONCEPT}}''. 
Regarding the concept scorer, the classifier treats the output sequence as the premise and generates the hypothesis for each concept by transforming it into the following format: ``This example is about {\textit{CONCEPT}}''. 
As for the baseline method, the classifier considers the question as the premise and generates the hypothesis for each concept by transforming it into the following format: ``This question is relevant to {\textit{CONCEPT}}''. 

\section{Ablation Studies of Hallucination Detection} \label{appendix:ablation}
\begin{figure*}[ht!]
    \centering
    % ROC Curve
    \begin{subfigure}[h]{0.3\textwidth}
        \centering
        \includegraphics[width=\textwidth]{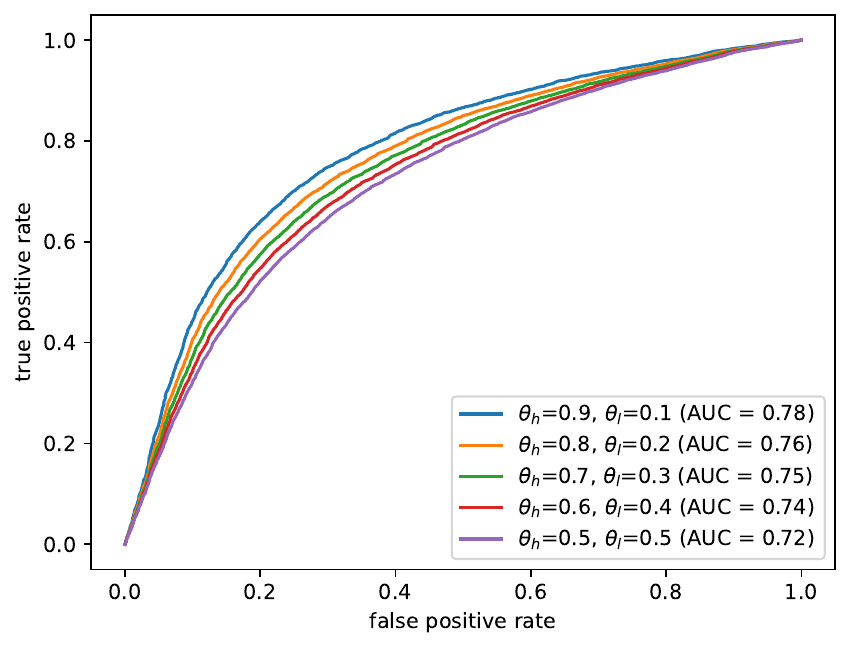}
        \caption{ROC Curve of Eli5-Category}
        \label{fig:eli5_roc}
    \end{subfigure}
    \hfill
    \begin{subfigure}[h]{0.3\textwidth}
        \centering
        \includegraphics[width=\textwidth]{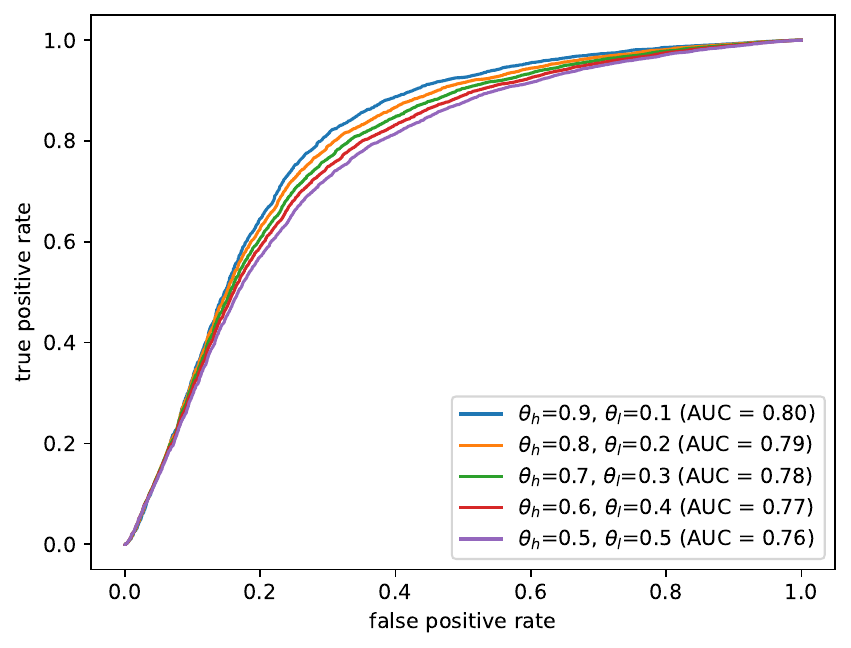}
        \caption{ROC Curve of Wiki-QA}
        \label{fig:wiki_roc}
    \end{subfigure}
    \hfill
    \begin{subfigure}[h]{0.3\textwidth}
        \centering
        \includegraphics[width=\textwidth]{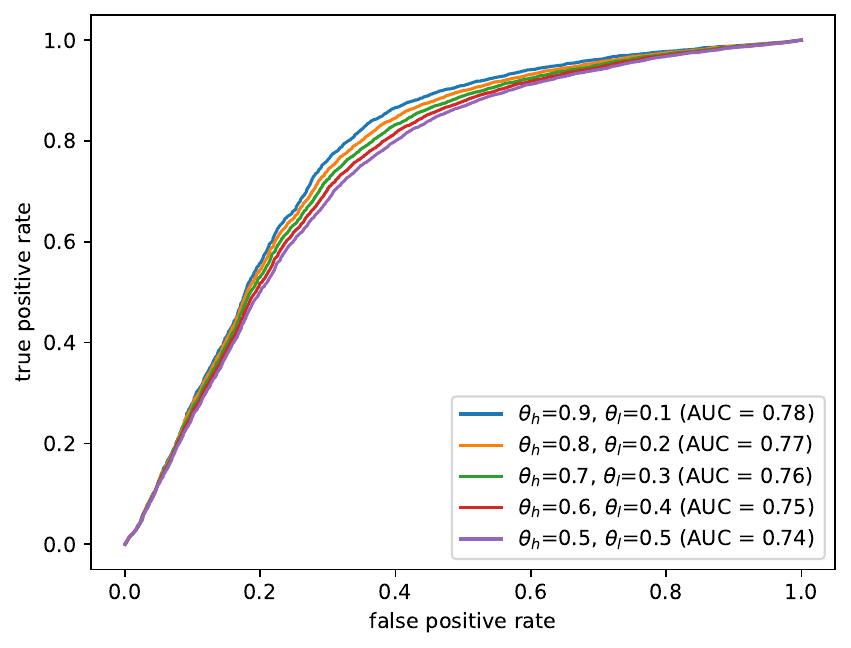}
        \caption{ROC Curve of QNLI}
        \label{fig:qnli_roc}
    \end{subfigure}
    \par\bigskip
    %  PR Curve
    \begin{subfigure}[h]{0.3\textwidth}
        \centering
        \includegraphics[width=\textwidth]{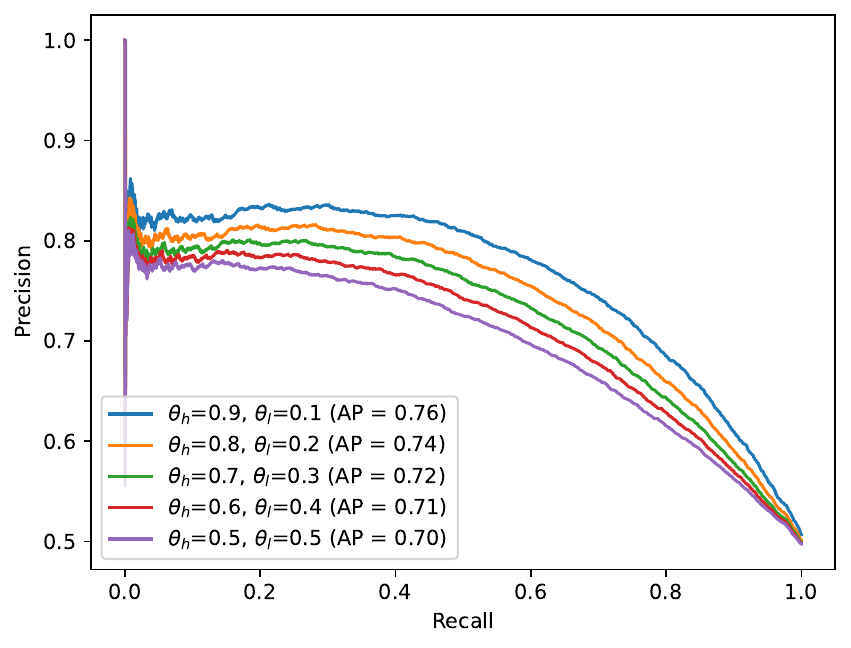}
        \caption{PR Curve of Eli5-Category}
        \label{fig:eli5_prc}
    \end{subfigure}
    \hfill
    \begin{subfigure}[h]{0.3\textwidth}
        \centering
        \includegraphics[width=\textwidth]{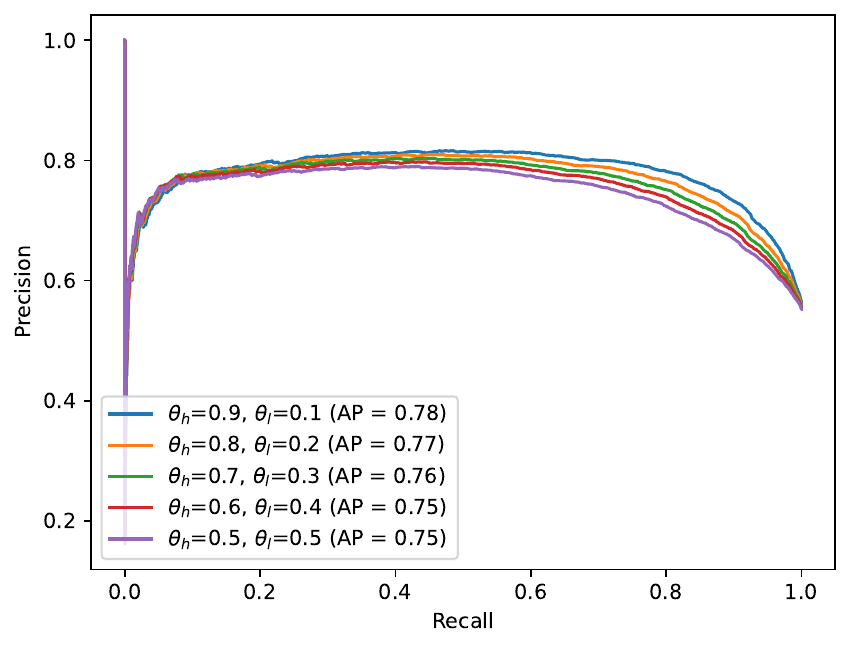}
        \caption{PR Curve of Wiki-QA}
        \label{fig:wiki_prc}
    \end{subfigure}
    \hfill
    \begin{subfigure}[h]{0.3\textwidth}
        \centering
        \includegraphics[width=\textwidth]{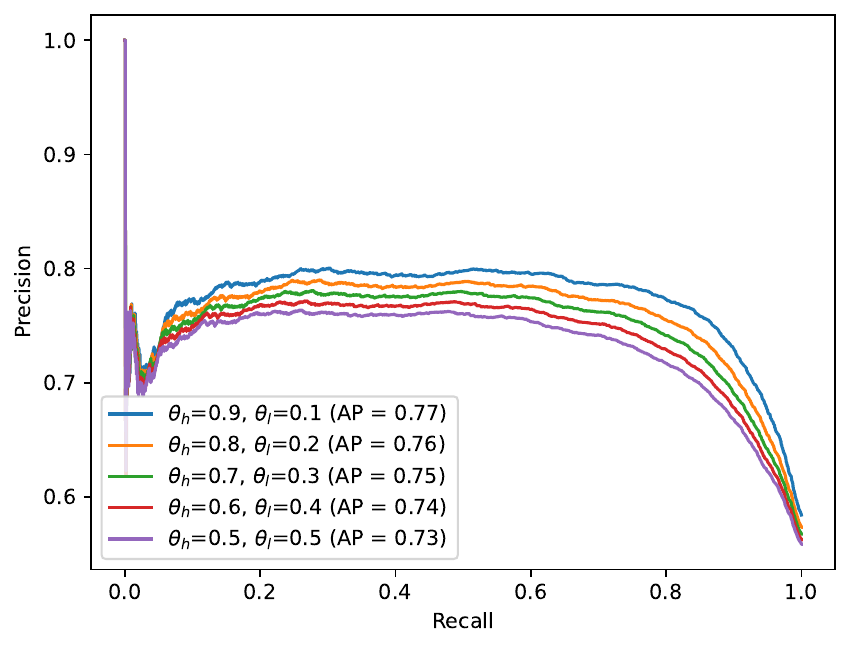}
        \caption{PR Curve of QNLI}
        \label{fig:qnli_prc}
    \end{subfigure}
    \caption{ROC Curves and PR Curves on different thresholds of concept score. The results indicate that our method demonstrates better performance when utilizing tighter thresholds. }
    \label{fig:roc_prc}
\end{figure*}
We further conduct ablation studies on the thresholds of concept scores. As illustrated in Figure~\ref{fig:roc_prc}, our method demonstrates better performance across all datasets when employing tighter thresholds -- specifically, a higher $\theta_{h}$  and a lower $\theta_{l}$. This observation implies that the scores predicted by our concept scorer effectively reflect the concept's faithfulness.

\section{Diversity Metric for Story Generation}
\begin{table*}[ht]
\centering
\begin{tabular}{c|c c c c c}
 & \multicolumn{5}{c}{\textbf{Number of stories in each tone}}\\
\textbf{Dataset} & \textbf{Happy} & \textbf{Sad} & \textbf{Humorous} & \textbf{Serious} & \textbf{Romantic}\\
\hline
Single-class dataset & 1000 & 0 & 0 & 0 & 0\\
Biased dataset & 600 & 100 & 100 & 100 & 100\\
Uniform distribution dataset & 200 & 200 & 200 & 200 & 200\\
\end{tabular}
\caption{Overview of story generation dataset.}
\label{tab:story_dataset}
\end{table*}

\begin{table*}[ht!]
\centering
\begin{tabular}{c|c c}
 & \multicolumn{2}{c}{\textbf{Diversity}}\\
\textbf{Dataset} & \textbf{Harmonic mean} & \textbf{Entropy} \\
\hline
Single-class dataset & 0.142 & 0.319 \\
Biased dataset & 0.903 & 1.286 \\
Uniform distribution dataset & 1.215 & 1.594\\
\end{tabular}
\caption{Results of our proposed diversity metrics. Both metrics successfully capture the diversity of ``tone'' across three datasets.}
\label{tab:diversity_metrics}
\end{table*}
\subsection{Related Work} \label{appendix:previous_diversity_metric}
Extensive research has leveraged LLMs for story generation tasks, and various metrics have also been introduced to evaluate the diversity of generated stories.
Existing metrics commonly rely on quantifying diversity through measures such as the count of distinct n-grams (\citealp{yao2019planandwrite}, \citealp{tevet-berant-2021-evaluating}, \citealp{li-etal-2016-diversity}, \citealp{goldfarb-tarrant-etal-2020-content}), or by employing BLEU or ROUGE scores (\citealp{papineni-etal-2002-bleu}, \citealp{zhu2018texygen}, \citealp{shu-etal-2019-generating}, \citealp{xie-etal-2023-next}, \citealp{tu-etal-2019-generating}). 
However, these metrics are confined to measuring lexical diversity and fail to capture high-level features such as tone or genre in story generation. 
While some diversity metrics based on text embeddings have been proposed to address this limitation (\citealp{lai-etal-2020-diversity}, \citealp{du-black-2019-boosting}), their applicability to story generation tasks remains unexplored. 

\subsection{Evaluation of Diversity Metric} \label{appendix:diversity_metric_evaluation}
To evaluate the effectiveness of our method as a diversity metric, we create three small datasets containing stories generated in different tones, as illustrated in Table~\ref{tab:story_dataset}. 
These datasets exhibit distinct distributions, with the highest expected diversity in the uniform distribution dataset and the lowest diversity in the single-class dataset. 
We utilize the prompt ``Generate a story in happy/sad/humorous/serious/romantic tone in five sentences.'' to generate the stories. 
The experiment results are presented in Table~\ref{tab:diversity_metrics}, demonstrating that the two proposed diversity metrics both effectively capture the diversity of the upper-level concept 'tone'.

\end{document}